\documentclass[10pt,twocolumn,letterpaper]{article}

\usepackage{cvpr}
\usepackage{times}
\usepackage{epsfig}
\usepackage{graphicx}
\usepackage{subfig}
\usepackage{amsmath}
\usepackage{amssymb}

\usepackage{algorithm}
\usepackage{algpseudocode}
\usepackage{listings}
\usepackage{breqn}
\usepackage{bm}
\usepackage{multirow}

\newcommand{\vv}{\bm v}
\newcommand{\hh}{\bm h}
\newcommand{\ww}{\bm w}
\newcommand{\WW}{\bm W}

\newcommand{\bb}{\bm b}
\newcommand{\cc}{\bm c}
\newcommand{\xx}{{\bm x}}

\newcommand{\F}{\mathcal{F}}
\DeclareMathOperator{\E}{\mathbf{E}}

\usepackage[breaklinks=true,bookmarks=false]{hyperref}

\cvprfinalcopy 

\def\cvprPaperID{****} 

\begin{document}

\title{Deep Restricted Boltzmann Networks}

\author{Hengyuan Hu \\
Carnegie Mellon University\\
{\tt\small hengyuanhu@cmu.edu}
\and
Lisheng Gao \thanks{Equal contribution.}\\
Carnegie Mellon University\\
{\tt\small lishengg@andrew.cmu.edu}
\and
Quanbin Ma \footnotemark[1]\\
Carnegie Mellon University\\
{\tt\small quanbinm@andrew.cmu.edu} \\
}

\maketitle

\begin{abstract}
Building a good generative model for image has long been an important topic in computer vision and machine learning. Restricted Boltzmann machine (RBM) \cite{rbm-hinton} is one of such models that is simple but powerful. However, its restricted form also has placed heavy constraints on the model's representation power and scalability. Many extensions have been invented based on RBM in order to produce deeper architectures with greater power. The most famous ones among them are deep belief network \cite{fast-learning-dbn}, which stacks multiple layer-wise pretrained RBMs to form a hybrid model, and deep Boltzmann machine \cite{dbm}, which allows connections between hidden units to form a multi-layer structure. In this paper, we present a new method to compose RBMs to form a multi-layer network style architecture and a training method that trains all layers/RBMs jointly. We call the resulted structure deep restricted Boltzmann network. We further explore the combination of convolutional RBM with the normal fully connected RBM, which is made trivial under our composition framework. Experiments show that our model can generate descent images and outperform the normal RBM significantly in terms of image quality and feature quality, without losing much efficiency for training.
\end{abstract}

\section{Introduction}
Boltzmann machine (BM) is a family of bidirectionally connected neural network models designed to learn unknown probabilistic distributions \cite{bm-original}. The original Boltzmann machine, however, is seldom useful as its lateral connections among both visible and hidden units make it computationally impossible to train. Restricted Boltzmann machine (RBM) \cite{rbm-hinton} is proposed to address this problem, where the connection pattern in a Boltzmann machine is restricted such that no lateral connections are allowed. This makes the learning procedure much more efficient while still maintains enough representation power to be a useful generative model \cite{rbm-collaborative-filtering}. Several deeper architectures are later invented to tackle the problem that one layer RBMs fail to model complicated probabilistic distributions in practice. Two of the most successful ones are deep belief network (DBN) \cite{fast-learning-dbn, reduce-dim-nn} and deep Boltzmann machine (DBM) \cite{dbm}. Deep belief network consists of multiple layers of RBMs trained in a greedy, layer-by-layer way. The resulted model is a hybrid generative model where only the top layer remains an undirected RBM while the rest become directed sigmoid belief network. Deep Boltzmann machine, on the other hand, can be viewed as a less-restricted RBM where connections between hidden units are allowed but restricted to form a multi-layer structure in which there is no intra-layer connection between hidden units. The resulted model is thus still a bipartite graph so that efficient learning can be conducted \cite{dbm, dbm-efficient-learning}. The learning procedure is often layer-wise pretrain followed by joint training of the entire DBM.

In this paper, we present a new way to compose RBMs to form a deep undirected architecture together with a learning algorithm that trains all layers jointly from scratch. We call the composed architecture deep restricted Boltzmann network (DRBN) because each layer consists of one RBM, and the semantic of our architecture is more similar to a multi-layer neural network than a deep Boltzmann machine. We also show that our model can be extended with convolutional RBMs for better scalability.

\section{Background}
In this section we will review restricted Boltzmann machine and its two major multi-layer extensions, \ie deep belief network and deep Boltzmann machine. Those three models are the foundations and inspirations of our new model. Therefore, it is crucial to understand them in order to identify the differences and advantages of our new deep restricted Boltzmann networks.

\begin{figure}[h!]
\centering   
\includegraphics[height=.08\textheight]{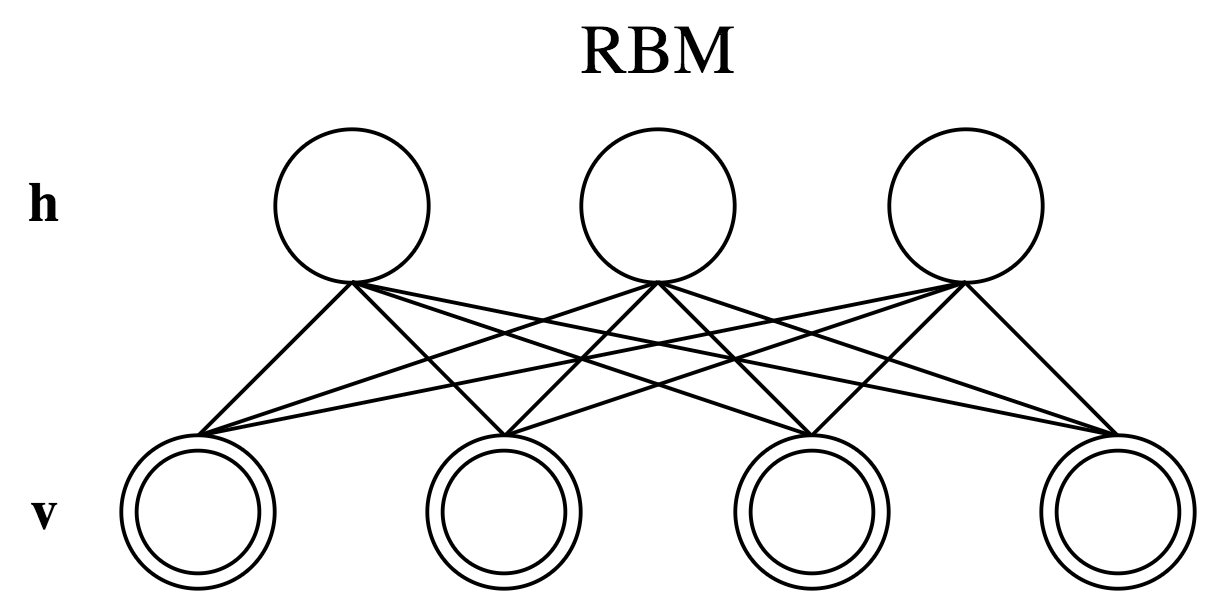} 
\caption{Restricted Boltzmann Machine.}
\label{fig:rbm}
\end{figure}

\subsection{Restricted Boltzmann Machine}
A restricted Boltzmann machine is an energy based model that can be viewed as a single layer undirected neural network. It contains a set of {\it visible units} $\vv \in \{0, 1\}^{D}$, {\it hidden units} $\hh \in \{0, 1\}^{P}$, where $D$ and $P$ are the numbers of visible and hidden units respectively. The parameters involved are $\theta=\{\WW,\bb,\cc\}$, denoting the mutual weights, visible units' biases and hidden units' biases. The {\it energy} of a given {\it state} $(\vv, \hh)$ is defined as:
\begin{align}
\E(\vv, \hh; \theta) &= -\bb^T\vv - \cc^T\hh - \vv^T\WW\hh \\
&= -\sum_{i=1} b_i v_i - \sum_{j=1} c_j h_j -\sum_{i,j} v_i W_{ij} h_j. \nonumber
\end{align}
The {\it probability} of a particular configuration of visible state $\vv$ in the model is
\begin{align}
p(\vv;\theta) &= {1 \over Z(\theta)} \sum_{\hh} e^{-\E(\vv,\hh;\theta)},   \label{vis-p} \\
Z(\theta) &= \sum_{\vv,\hh} e^{-\E(\vv,\hh;\theta)},
\end{align}
where $Z(\theta)$ is the partition function. Because RBM restricts the connections in the graph such that there is no link among visible units or among hidden units, the hidden units $p_j$ become conditionally independent given the visible state $\vv$, and vice versa. Hence, the conditional probability of a unit has the following simple form:
\begin{align}
p(h_{j} = 1|\vv) &= \sigma \left(\sum_{i\vphantom{j}} v_i W_{ij} + c_j \right), \label{eq:compute-up}\\
p(v_{i} = 1|\hh) &= \sigma \left(\sum_{j} h_j W_{ij} + b_i \right), \label{eq:compute-down}
\end{align}
where $\sigma(x) = 1 / (1 + \exp(-x))$ is the sigmoid function. This property allows efficient parallel block Gibbs sampling alternating between $\vv$ and $\hh$, and thus makes the learning process faster.

The learning algorithm for RBM is conceptually simple. With the probability of visible state defined in Equation~\ref{vis-p}, we can perform gradient descent to maximize $p(\vv)$. The update rule for parameters can then be derived by computing the derivative of the negative log-likelihood function with regard to each parameter. However, in order to facilitate later discussion, we take a detour to derive the learning rule under a more general energy based model (EBM) framework.

First we define the {\it free energy} of a visible state $\vv$ as
\begin{align}
\F(\vv) &= -\log \sum_{\hh} e^{-\E(\vv,\hh)}.
\end{align}
Although this is a sum of exponential amount of terms, it can be easily computed for RBM with binary hidden units. It can be proved that $\F(\vv)$ can be rewritten as
\begin{align}
\F(\vv) &= -\sum_i  b_i v_i - \sum_j \log \sum_{h_j} e^{h_j(c_i+\sum_i v_i W_{ij})}.
\end{align}
With free energy, the probability of a given visible state in Equation~\ref{vis-p} can be simplified as
\begin{align}
p(\vv; \theta) &= {1 \over Z} e^{-\F(\vv)} = {e^{-\F(\vv)} \over \sum_\xx e^{-\F(\xx)}}.
\end{align}
We then derive the derivatives as 
\begin{align}
{\partial -\log p(\vv^{(i)}) \over \partial \theta} = {\partial \F(\vv^{(i)}) \over \partial \theta} - \sum_j p(\vv^{(j)}) {\partial \F(\vv^{(j)}) \over \partial \theta} \label{eq:gradient1}
\end{align}
where $\vv^{(i)}$s are the current training data,  $\vv^{(j)}$s are all the possible outputs of visible units that can be generated by the model.

\begin{algorithm}[t]
\caption{PCD(k, N)}
\label{algo:pcd}
\begin{algorithmic}[1]
\State Randomly initialize $N$ particles $\vv^{(1)}_0, \vv^{(2)}_0, \hdots, \vv^{(N)}_0$.
\For{$t=1$ to NUM\_ITERATION}
	\ForAll{$\vv^{(j)}_{t}, j=1,2,\hdots,N$}
		\State Do $k$ Gibbs sampling iterations to get $\vv^{(j)}_{t, k}$.
	\EndFor
   	\State $\vv^{(j)}_{t+1} \gets \vv^{(j)}_{t, k}$.
    \State Use $\vv^{(j)}_{t+1}$s and Eq.~\ref{eq:gradient2} to compute gradients.
    \State Update parameters with the gradients.
\EndFor
\end{algorithmic}
\end{algorithm}

While the first term in the equation, often noted as the {\it data-dependent term}, can be computed directly given training data, the second term, often noted as the {\it model-dependent term}, is almost impossible to compute as the number of possible $\vv^{(j)}$s is exponential to the input size. Persistent contrastive divergence (PCD) \cite{pcd,Neal-pcd} has been widely employed to estimate the second term. The algorithm works as shown in Algorithm~\ref{algo:pcd}, where $N$, the chain size, denotes the number of PCD particles used. Using PCD algorithm to approximate the model-dependent term, Equation~\ref{eq:gradient1} becomes
\begin{align}
{\partial -\log p(\vv^{(i)}) \over \partial \theta} = {\partial \F(\vv^{(i)}) \over \partial \theta} - \frac{1}{N} \sum_j {\partial \F(\vv^{(j)}) \over \partial \theta}. \label{eq:gradient2}
\end{align}
This will be a key equation for parameter updates in later sections.


\begin{figure}[t!]
\centering
\subfloat[][]{\includegraphics[height=.15\textheight]{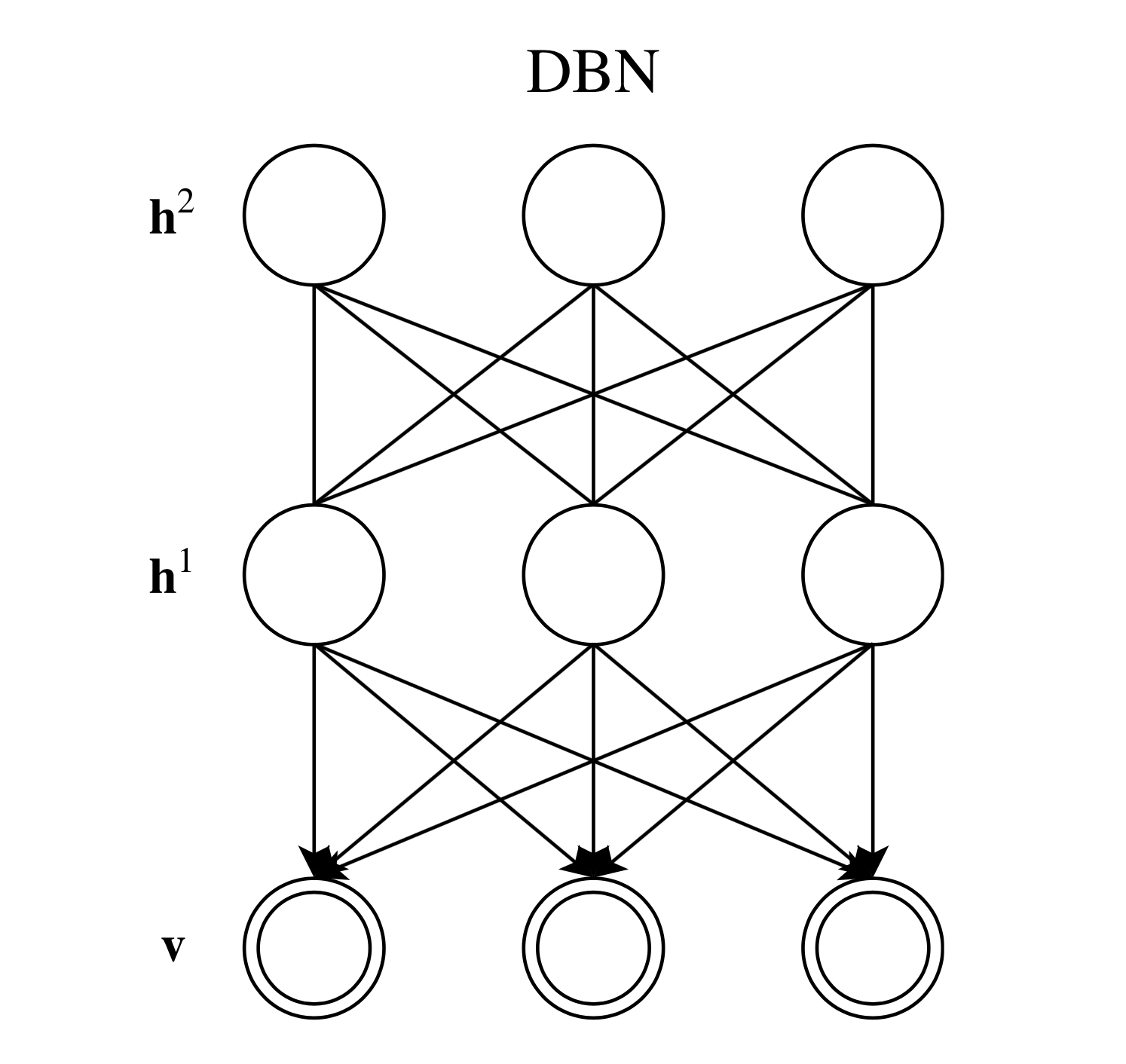} \label{fig:dbn}} \quad
\subfloat[][]{\includegraphics[height=.15\textheight]{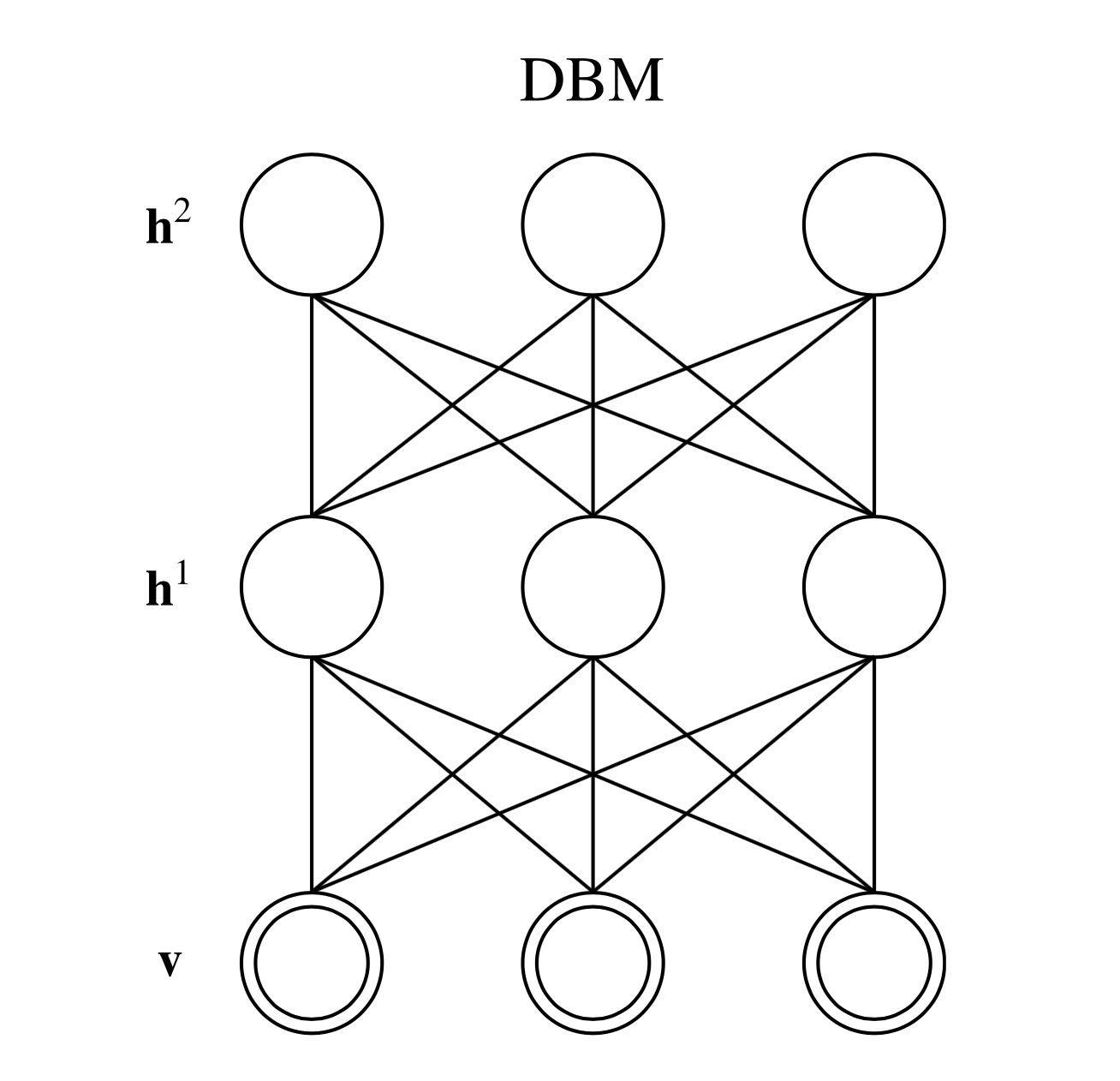} \label{fig:dbm}} 
\caption{Architectures of {\bf (a)} Deep Belief Network; {\bf (b)} Deep Boltzmann Machine.}
\label{fig:rbms}
\end{figure}

\subsection{Deep Belief Network}
Deep belief network, as shown in Figure~\ref{fig:dbn}, is a deep architecture built upon RBM to increase its representation power by increasing depth. In a DBN, two adjacent layers are connected in the same way as in RBM. The network is trained in a greedy, layer-by-layer manner \cite{fast-learning-dbn}, where the bottom layer is trained alone as an RBM, and then fixed to train the next layer. After all layers are pretrained, the resulted network contains one RBM at the top layer while the rest layers form a directed neural network. There is no joint training for all layers in DBN. To generate images, we need to first run Gibbs sampling in the top RBM till convergence and then propagate it down to the bottom layer. DBN has also been extended to use convolutional RBMs for better scalability and higher quality features \cite{conv-dbn}.

\subsection{Deep Boltzmann Machine}
Deep Boltzmann machine \cite{dbm}, as shown in Figure~\ref{fig:dbm}, is another special model in the Boltzmann machine family. Different from RBM which allows no connections between hidden units, DBM introduces additional connections between latent variables to form a multi-layer structure in the hidden part of Boltzmann machine. The energy of a DBM is defined as
\begin{align}
\E(\vv, \hh^{1}, \hh^{2}; \theta) &= -\bb^T\vv - \cc^{(1)T}\hh^{(1)} - \cc^{(2)T}\hh^{(2)} \nonumber \\
&- \vv^T\WW^{(1)}\hh^{(1)} - \hh^{(1)T}\WW^{(2)}\hh^{(2)}.
\end{align}
Given the energy function, the conditional distributions of each layers can be derived as:
\begin{align}
p(h^{(2)}_{j} = 1|\hh^{(1)}) &= \sigma \left( \sum_{i} h^{(1)}_i W^{(2)}_{ij} + c^{(2)}_j \right) , \label{eq:dbm-compute-up} \\
p(h^{(1)}_{j} = 1|\hh^{(2)}, \vv) &= \sigma \left(\sum_{i} h^{(2)}_i W^{(2)}_{j,i} \right. \nonumber \\
&\hphantom{=\sigma} \left. + \sum_{k} v_k W^{(1)}_{k,j} + c^{(1)}_j \right), \label{eq:dbm-compute-down-1}\\
p(v_{i} = 1|\hh^{(1)}) &= \sigma \left( \sum_{j} h^{(1)}_j W^{(1)}_{ij} + b^{(1)}_i \right) . \label{eq:dbm-compute-down-2}
\end{align}
Note that the equation for the middle layer is different because it depends on both its adjacent neighbors. Block Gibbs sampling can still be performed by alternating between odd and even layers, which makes efficient learning possible. Further more, mean-field method and persistent contrastive divergence \cite{pcd, Neal-pcd} are employed to make the learning tractable \cite{dbm,dbm-efficient-learning}. Note that DBM also needs greedy layer-wise pretraining to reach its best performance when the number of hidden layers is greater than 2.

\section{Deep Restricted Boltzmann Network}
Both deep belief network and deep Boltzmann machine are rich models with enhanced representation power over the simplest RBM but more tractable learning rule over the original BM. However, it is interesting to see whether we can devise a new rule to stack the simplest RBMs together such that the resulted model can both generate better images and extract higher quality features. In this section, we will introduce the architecture of deep restricted Boltzmann network and its training method. In addition, we will further extend our architecture to support both RBMs and convolutional RBMs.

\subsection{Architecture}
\begin{figure}[t!]
\centering 
\subfloat[][]{\includegraphics[height=.15\textheight]{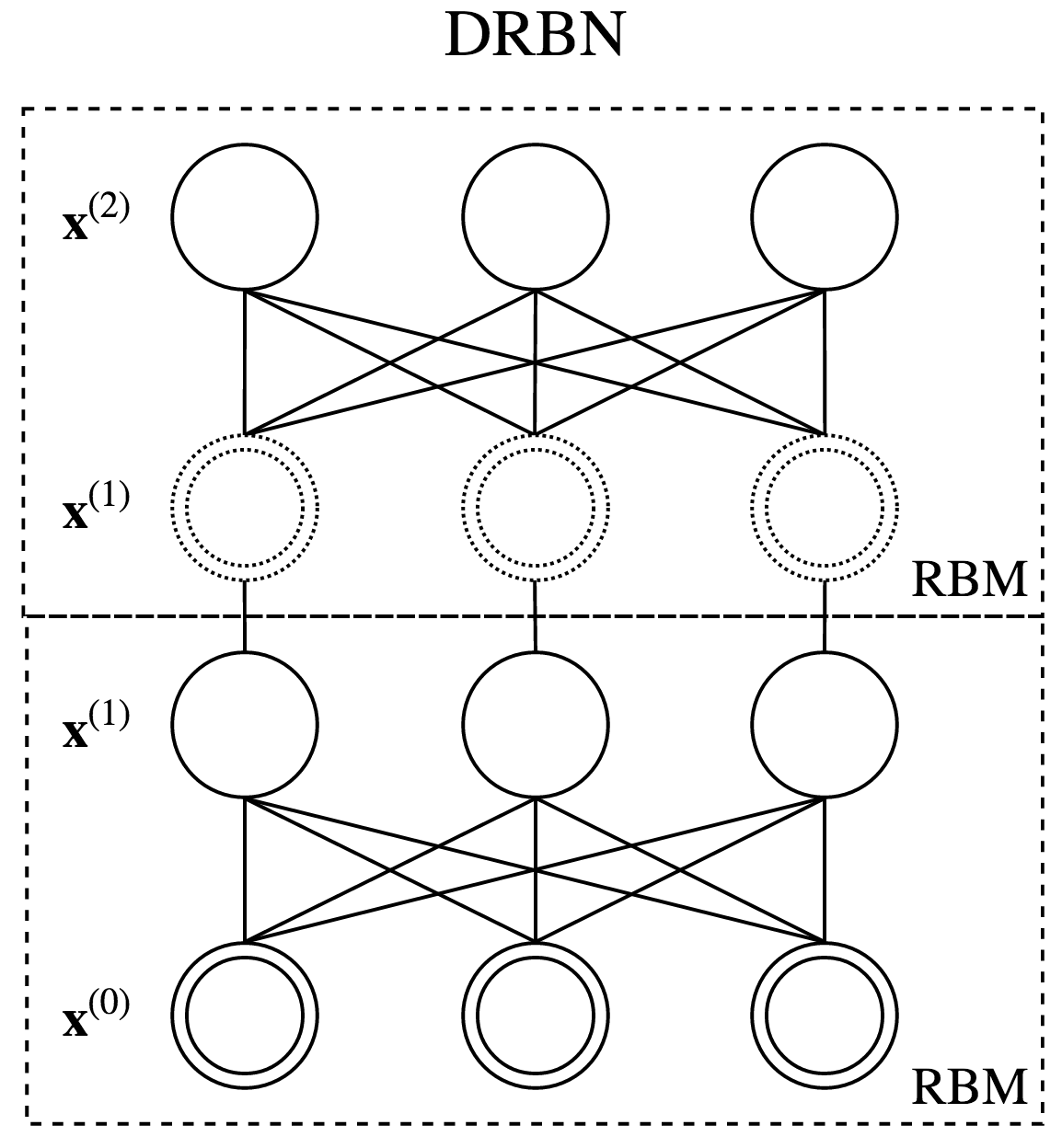} \label{fig:drbn_arch}} \quad
\subfloat[][]{\includegraphics[height=.15\textheight]{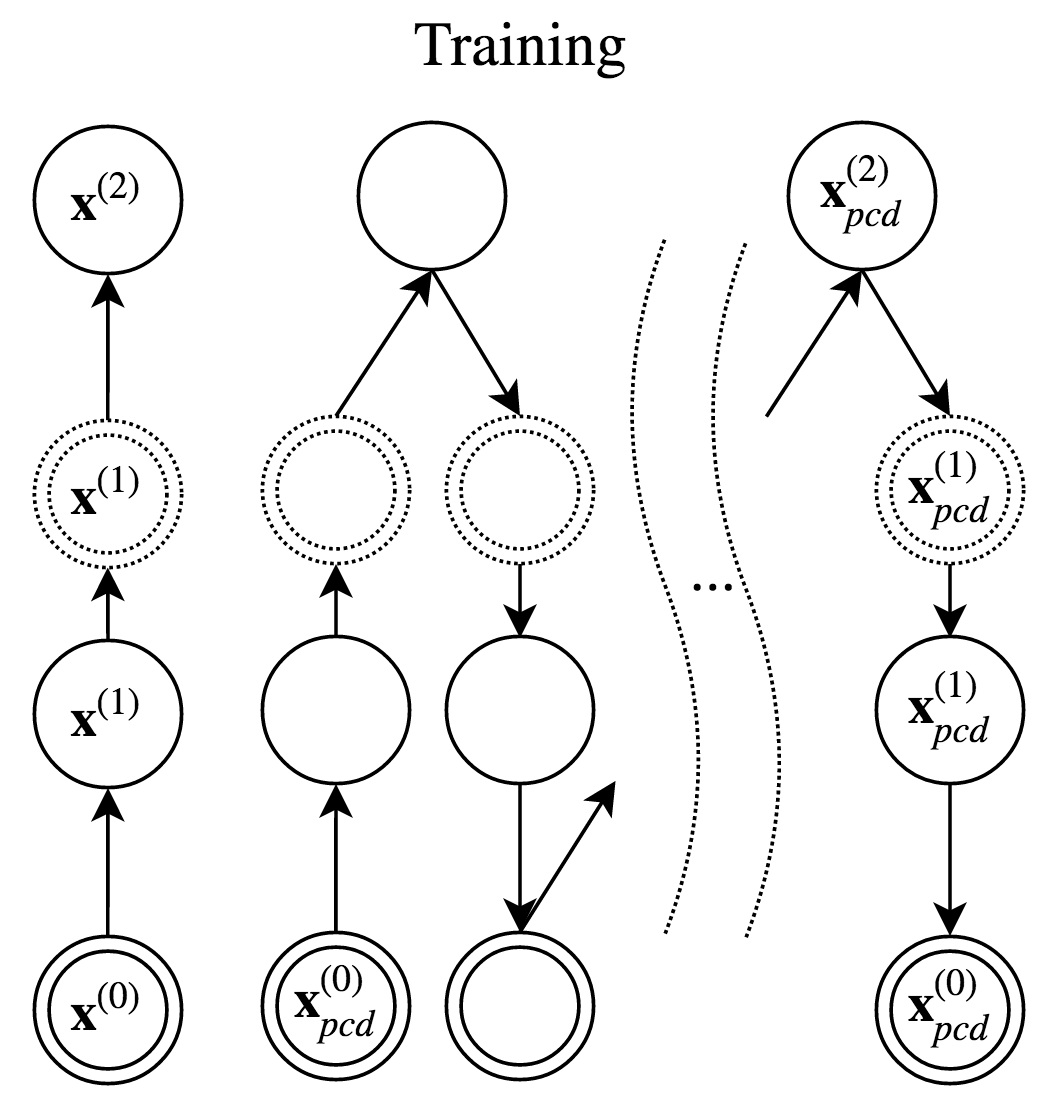} \label{fig:drbn_train}}

\caption{{\bf (a)} Architecture of Deep Restricted Boltzmann Network; {\bf (b)} Illustration of the learning process using PCD.}
\label{fig:drbn}
\end{figure}

Deep restricted Boltzmann network is a multi-layer neural network where each layer is a strictly restricted Boltzmann machine. As shown in Figure~\ref{fig:drbn_arch}, each RBM's hidden units pass their values directly to the visible units of next layer. Now, hidden units at each layer are in fact also the visible units in the next layer, so we will not differentiate between $\vv$ and $\hh$, while only use $\xx^{(l)}$ to denote the state of the $l$-th layer for simplicity of notation. In other words, $\hh^{(i)}$ and $\vv^{(i+1)}$ are unified as $\xx^{(i+1)}$ in the following discussion because they essentially hold the same value. Naturally, $\xx^{(0)}$ denotes the input layer of the entire network. Because each layer is viewed as a standalone RBM instead of part of a DBM, the energy is thus defined on each layer, \ie each RBM, instead of the entire network. Therefore, the $l$-th layer, we have:
\begin{align}
 &\E(\xx^{(l)}, \xx^{(l+1)}; \theta^{(l)}) \nonumber \\
=& -\bb^{(l)T}\xx^{(l)} - \cc^{(l)T}\xx^{(l+1)} - \xx^{(l)T}\WW^{(l)}\xx^{(l+1)}.
\end{align}
In addition, the computation of probabilities for each layer during the upward and downward pass is exactly the same as that in RBM:
\begin{align}
p(x^{(l+1)}_{j} = 1|\xx^{(l)}) &= \sigma \left(\sum_{i} x^{(l)}_i W^{(l)}_{ij} + c^{(l)}_j \right) \label{deep-up} ,\\
p(x^{(l)}_{i} = 1|\xx^{(l+1)}) &= \sigma \left(\sum_{j} x^{(l)}_j W^{(l)}_{ij} + b^{(l)}_i \right). \label{deep-down}
\end{align}
One full iteration of Gibbs sampling in a deep restricted Boltzmann network is illustrated as one up-and-down cycle in Figure~\ref{fig:drbn_train}. It is similar to a forward-backward pass happens in the neural network. Given the input visible state $\xx^{(0)}$, we compute upward layer-by-layer using Equation~\ref{deep-up} until we reach the top $\xx^{(L)}$. Then we compute downward following Equation~\ref{deep-down} to obtain a new sample. From the sampling process, we can also clearly see the difference between our model and existing deep models such as DBN and DBM.

\subsection{Training DRBN}
All RBMs in the network are trained jointly with persistent contrastive divergence. Starting with input visible $\xx^{(0)}$, the model would first do a upward pass following Equation~\ref{deep-up} with Bernoulli sampling at each layer. The results obtained in this pass are denoted as $\{\xx^{(l)} | l = 0, 1, 2, ..., L\}$, and will later be used to calculate the data-dependent term. Then $k$ upward-downward passes are performed on the PCD particles. The values of these particles at each layer during the last downward pass are obtained as $\{\xx^{(l)}_{pcd} | l = 0, 1, 2, ..., L\}$\footnote {Note that this diverges from our previous notation for RBM, where the superscript now identifies the layer instead of the particle.}, which are then used to approximate the model-dependent term. These two processes are illustrated in Figure~\ref{fig:drbn_train}. After obtaining $\xx^{(l)}$s and $\xx^{(l)}_{pcd}$s, we can compute the gradient for each layer separately using the approximation in Equation~\ref{eq:gradient2}. Parameters $\theta^{(l)}$  of all layers are then updated simultaneously.

This learning rule further enforces our assumption that each RBM in the network are treated separately since all the information required to compute free energy $\F$ and gradients is contained locally in every RBM. An intuition behind such training method is that each layer tries to decrease the free energy of training examples while increase the free energy of the invalid examples generated at its own level.

\begin{figure*}[!th]
\centering
\subfloat[][]{\includegraphics[width=.23\textwidth]{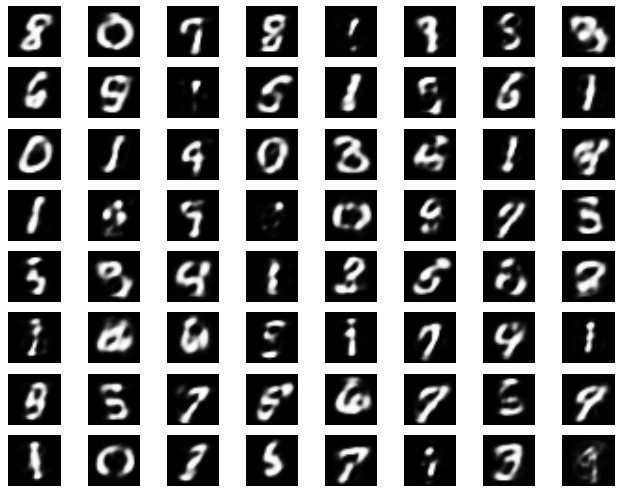}} \quad
\subfloat[][]{\includegraphics[width=.23\textwidth]{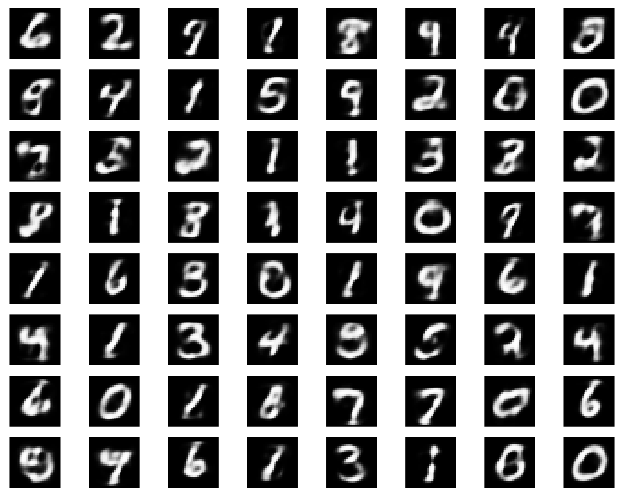}} \quad
\subfloat[][]{\includegraphics[width=.23\textwidth]{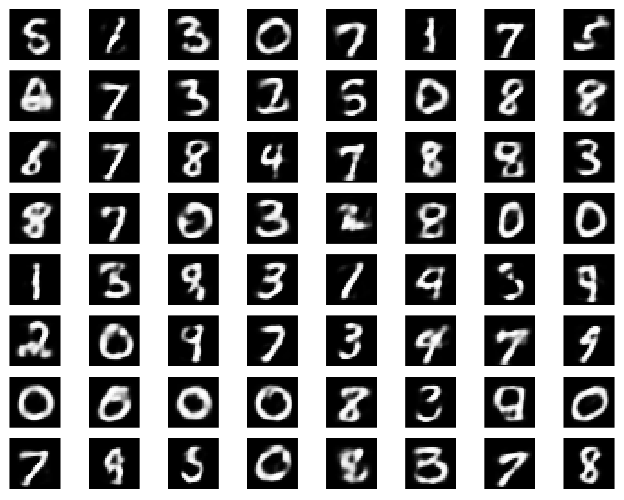}} \quad
\subfloat[][]{\includegraphics[width=.23\textwidth]{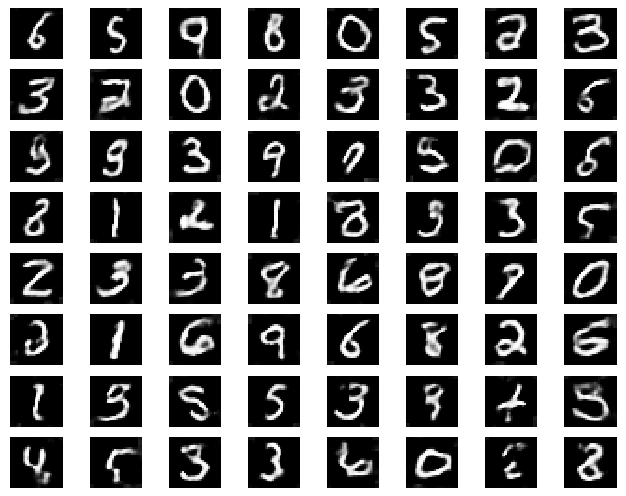}}

\caption{Randomly drawn samples from \{ {\bf (a)} images generated by RBM with 1000 hidden units; {\bf (b)} images generated by DRBN with two hidden layers of size 500 and 1000; {\bf (c)} images generated by DRBN with three hidden layers of size 500, 500 and 1000; {\bf (d)} images generated by the discussed convolutional DRBN \}.}
\label{fig:mnist}
\end{figure*}

\subsection{Extension with Convolutional RBM}
\label{sec:CRBM}
The original RBM operates in a fully connected way, which means that it takes the entire input image as a one dimensional array and ignores locality. Inspired by the huge success of convolutional neural network in computer vision \cite{alexnet,resnet}, we extend our method to utilize convolutional RBM under the same composition framework. Convolutional RBM \cite{conv-dbn,shapebm} is conceptually the same as RBM, but shares weights (filters) among all location on the input image and replaces the matrix multiplications happen during the upward-downward computation with convolutions. For notation simplicity, let's assume the visible units are of shape $(N_\vv, N_\vv)$ with 1 channel, the hidden units are of the shape $(N_\hh, N_\hh)$ with $C$ channels, and the convolution uses filter of size $(N_\ww, N_\ww)$ with stride 1 and no padding. Similar formulae can be derived for all other valid settings for convolution in the same way. We use $f(\xx, \WW)$ and $f^T(\xx, \WW)$ to denote convolution and transposed convolution between input $\xx$ and filters $\WW$. The energy $\E(\vv, \hh)$ can then be defined as
\begin{align}
\E(\vv, \hh) = &- \sum_{k=1}^{C} \sum_{i,j = 1}^{N_h} \sum_{r,s = 1}^{N_w} h_{i,j}^k W_{r,s}^k v_{i+r-1, j+s-1} \nonumber \\
&- \sum_{k=1}^{C}c_k \sum_{i,j=1}^{N_h}h_{i,j}^{k} - b\sum_{i,j=1}^{N_v} v_{i,j}.
\end{align}

Similar to Equation~\ref{eq:compute-up} and \ref{eq:compute-down}, the conditional probabilities that are used for block Gibbs sampling are
\begin{align}
p(h_{ij}^{k} = 1|\vv) &= \sigma \left(f(\vv, W)_{i,j}^{k} + c_k \right),\\
p(v_{i} = 1|\hh) &= \sigma \left(f^T(\hh, W)_{i,j} + b \right).
\end{align}
And the free energy for convolutional RBM is
\begin{align}
\F(\vv) &= -\log \sum_{\hh} e^{-\E(\vv,\hh)} \\
&= -b\sum_{i,j} v_{ij} - \sum_{k,i,j}\log \left( 1+e^{\alpha_{ij}^k} \right), \\
\alpha^k_{ij} &= \sum_{r,s}^{N_w} W_{r,s} v_{r+i-1,s+j-1} + c^k.
\end{align}
With the formula of free energy, the gradient can be computed easily by plugging this new $\F$ into Equation~\ref{eq:gradient2}.

Previously, convolutional RBM has been studied in \cite{conv-dbn,stack-conv-rbm}, but with the focus on feature extraction instead of image generation. Empirically, we find that unlike RBM, convolutional RBM itself cannot generate meaningful images after trained on dataset like MNIST. This may be caused by the fact that one layer convolution handles only local information and lack connections between different receptive fields and feature channels. Therefore the generated images have no global coherence. With our composition framework, however, several convolutional RBMs and normal RMBs can be connected together to form a convolutional deep restricted Boltzmann network, just like using convolution layers and fully connected layers to form a convolutional neural network. The resulted architecture, as later shown in the experiment section, can generate reasonable images.

\section{Experiments}
We implement our model in Tensorflow \cite{tensorflow2015-whitepaper}. With the help of its auto-differentiation functionality, we can simply define the loss function for each RBM as
\begin{align}
Loss(\theta^{(l)}) = \frac{1}{M}\sum_{i}\F(\xx^{(l)}_{i}) - \frac{1}{N}\sum_{j}\F(\xx^{(l)}_{j})
\end{align} 
and let the auto-diff and gradient based optimizers take care of the rest. The $M$, $N$ in the previous equation denote the size of the minibatch and the number of PCD particles, respectively. In all the following experiments, we run PCD for 5 iterations between each update and set the number of PCD particles to 100, \ie $PCD(5, 100)$ in Algorithm \ref{algo:pcd}. The generated images are the {\it probabilities} of the binary visible units, \ie no sampling are performed in the first layer during the last downward pass. The code to reproduce our experiment results will be released.

\subsection{MNIST}
The MNIST digit dataset contains 50,000 training, 10,000 validation and
10,000 test images of ten handwritten digits (0 to 9), each with a resolution of $28\times 28$ pixels. To test the capacity of deep restricted Boltzmann networks in generating images, we trained two fully-connected DRBN and an experimental convolutional version. We compare the images generated with a plain RBM. We also tested these models in a semi-supervised context.

The structures of the two fully-connected DBRN are as follows: one with two hidden layers (500 and 1000 hidden units; 0.9 million parameters), and the other with three hidden layers (500, 500 and 1000 hidden units; 1.15 million parameters). During the training process, all of 50,000 training images of MNIST are used and both the size of minibatch and size of PCD particles are set to 100. The initial states are uniform random noise and we run Gibbs sampler 10,000 times to produce the output. 

The results are compared with the images generated by RBM with a hidden layer of size 1000. Figure~\ref{fig:mnist} shows the randomly-selected generated images by the three models respectively. We can observe that all of them are capable of generating rather recognizable digits. The images generated by plain RBM are noisy, blurred at boundaries and occasionally broken. The results of DRBN, however, have much less noise, sharper boundaries and smoother strokes. Furthermore, 3-layer model performs better in this case in terms of image quality than its shallower alternative. 

We also explore the possibilities of generalizing the architecture to support convolution layers as discussed in Section~\ref{sec:CRBM}. It is worth noting that during our experiments, several key settings have been critical for generating reasonable results. First, a rather large filter size has to be chosen for the first convolution layer, otherwise the generated digits will look scattered and have extremely bumpy edges. Second, we fail to produce any meaningful result if a stride of 1 is used. The reason would be that if the filters are densely applied, each visible unit will receive inputs from many filters during the downward computation and those values are hard to coordinate, making the model difficult to train. That being said, it is still essential for the filters to overlap each other, otherwise the generated images will be discontinuous. These properties are different from those of convolution layers in discriminative models \cite{vgg, resnet} where performance generally benefits from smaller and denser filters. Third, as mentioned in Section \ref{sec:CRBM}, the fully-connected layer at the top plays an important role and distinguishes our model from traditional convolutional RBM, which can hardly generate any meaningful images.

Under such design guidelines, we will use a three-layer convolutional DRBM for subsequent comparisons. The first layer consists of $64$ filters of size $12 \times 12$, and the second layer has $128$ filters of size $5 \times 5$. Both layers use a stride of 2. The outputs units are then flattened and connected to an RBM with $512$ units. In total the model uses around 0.6 million parameters. The generated images are also shown in Figure~\ref{fig:mnist}. 

These digits look indeed legit, which alone is much better than traditional convolutional RBM. Arguably, these generated digits are not as clear and coherent as fully-connected ones, which is probably due to the loss of information through the process of convolution. When it comes to generative tasks, the locational invariance of convolution does not come in so handy as they were in classification and feature extraction, especially during the reconstruction phase, where the relative locality actually matters. 

It is common for an image dataset to have only a small proportion manually labeled, while the remaining majority of raw data are of unknown categories. Therefore, it is worth exploring the performance of our model under such semi-supervised learning scenario to see whether the DRBN and convolutional DRBN can extract useful features. To convert our models for discriminative task, we add a fully connected layer with ten neurons using softmax activation to the pretrained model, mapping the values in the top hidden units in the original DRBN to a ten-dimensional vector. We evaluate the performance of different setups of DRBNs and compare their results with RBM, together with a plain fully-connected classifier trained from the scratch using the limited training data.

We carry out the experiments in two different phases. In the first phase, we keep the previously learnt DRBN parameters fixed while only training parameters in the new softmax layer. The purpose is to compare the quality of features extracted by different architectures and verify that a deeper model can indeed extract better features due to multiple layers of hierarchy. Table~\ref{table:mnist_semi} shows the results of different models trained on 600, 3000 and 6000 images (1/10 of the set used for previous unsupervised learning) randomly drawn from the dataset, and evaluated on the entire test set containing 10,000 images. We used Adam \cite{adam} with the default parameters to perform gradient descent. We can conclude that both RBM and DRBN can extract useful features that help the classification when labeled data is scarce, and our DRBN models outperform normal RBM by a large margin. As expected, our convolutional DRBN prove to work better than fully-connected ones in this case, as we only need to do a one-way upward pass for feature extraction.

\begin{table}[t]
\begin{center}
\begin{tabular}{|l|c|c|c|}
\hline
\multirow{2}{*}{Model} &
	\multicolumn{3}{c|}{Test Error} \\
		& 0.6K & 3K & 6K \\
\hline\hline
Plain FC & 13.31\% & 9.46\% & 8.89\% \\
RBM & 9.15\% & 5.81\% & 5.25\% \\
3-layer DRBN & 7.45\% & 4.84\% & 4.67\% \\
Conv DRBN & 6.94\% & 3.89\% & 3.43\% \\

\hline
\end{tabular}
\end{center}
\caption{Test error after training only the last softmax layer with DRBN parameters frozen, using randomly-drawn labeled training data of size 600, 3000 and 6000, respectively. Each value shown was taken average over 10 independent runs.}
\label{table:mnist_semi}
\end{table}

To further examine the quality of features extracted by DRBN, we take the trained models in the previous phase and fine tune all the layers together using the same amount of training data but with a smaller learning rate, to see how much increase we can get by allowing the feature extractors to adjust themselves. As shown in Table~\ref{table:mnist_semi_flex}, the increment in the performance is quite small, especially when the data is rare. This proves that the original DRBNs have already extracted nearly optimal features, which makes further fine-tuning redundant when the data is scarce.

The numbers seem to still have plenty of space for improvement, but it is important to keep in mind that they are obtained under the assumption of scarce labeled data. As we can imagine, if we supply the whole 60,000 MNIST training and validation set to the network, we could definitely get a better result, but then we would be just training a traditional neural network.

\begin{table}[t]
\begin{center}
\begin{tabular}{|l|c|c|c|}
\hline
\multirow{2}{*}{Model} &
	\multicolumn{3}{c|}{Test Error} \\
		& 0.6K & 3K & 6K \\
\hline\hline
3-layer DRBN & 7.26\% &  3.83\% &  3.29\% \\
Conv DRBN & 6.90\% & 3.52\% & 2.92\% \\

\hline
\end{tabular}
\end{center}
\caption{Test error after fine-tuning the whole network from last phase. Each value shown was taken average over 10 independent runs.}
\label{table:mnist_semi_flex}
\end{table}

\subsection{Weizmann Horses}

\begin{figure*}[!th]
\centering
\subfloat[][]{\includegraphics[width=.3\textwidth]{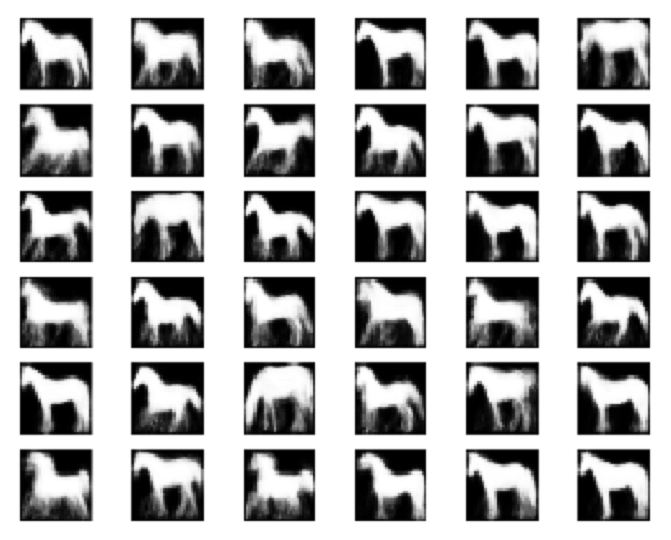}} \quad
\subfloat[][]{\includegraphics[width=.3\textwidth]{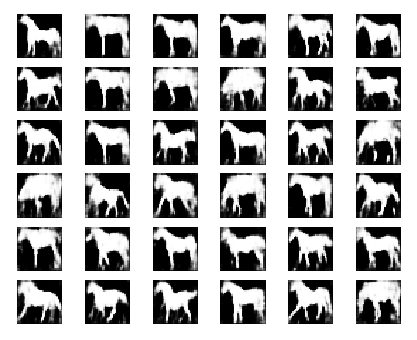}} \quad
\subfloat[][]{\includegraphics[width=.3\textwidth]{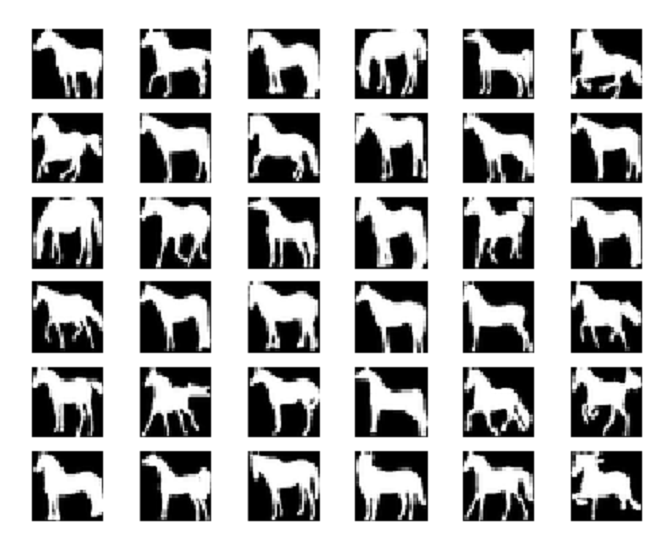} \label{fig:weizmann_original}}

\caption{Randomly drawn samples from \{ {\bf (a)} images generated by DRBN with two hidden layers of size 500 and 1000; {\bf (b)} images generated by the discussed convolutional DRBN; {\bf (c)} pre-processed training images \}. }
\label{fig:weizmann}
\end{figure*}

We further test our model on the Weizmann Horse dataset \cite{weizmann}, which includes 328 binary horse figures segmented from real world photos, with quite a variety of different postures. The images also have a high variance especially due to its abundant details regarding the necks, feet and tails of horses under the states of resting, running and eating, as will be shown later in Figure~\ref{fig:weizmann_original}. The high variety and low volume of the dataset make it a challenging task for the model to generate realistic horses with various postures. The original images come in varying resolutions, so we crop the horses to the center and reshape them to $32 \times 32$ pixels. Now the dataset resembles the taste of MNIST, only much smaller.

Similar to the experiments in MNIST, We build a fully-connected DRBN and a convolutional DBRN to learn from this dataset, and then generate horse images using Gibbs sampling with random noise initialization. For the fully-connected model, we choose a 2-layer structure with the number of hidden units being 500 and 1000 respectively. For the convolutional model, we keep the same structure we have used for MNIST. We train these two model using all pre-processed images. The results are shown in Figure~\ref{fig:weizmann} together with the original dataset. Those generated by fully-connected DRBN tend to have better clarity and smoother strokes along the boundaries, but the image would look blurred as a whole. Sometimes we could also identify two generated images that look very alike. Those generated by the convolutional DRBN, on the other hand, have shown greater variety of general postures and details in legs, though at the price of more noise.

Again, for the convolutional DRBN, we cannot generate descent images if the first layer uses a filter as small as $5 \times 5$. Such pattern for model selection is also reflected in ShapeBM \cite{shapebm}, whose implementation of deep Boltzmann machine on similar $32 \times 32$ horse images is equivalent to using 500 filters of size $18 \times 18$ with stride $14$ in the first layer. This agrees with our analysis above that such locality has to be captured, and later combined by the fully-connected layers for a successful reconstruction.

\begin{figure}[h]
\centering
\includegraphics[width=\linewidth]{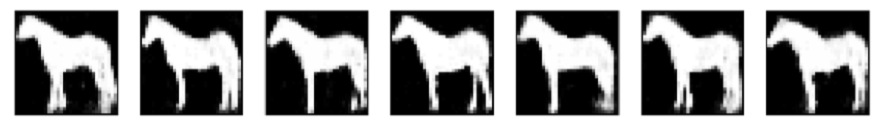} \\
\includegraphics[width=\linewidth]{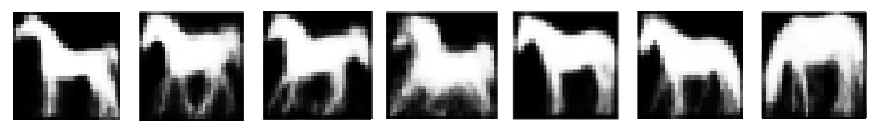} \\
\includegraphics[width=\linewidth]{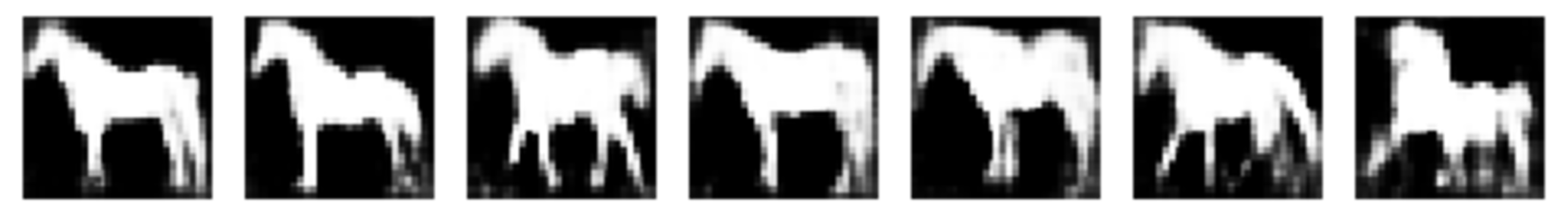}
\caption{{\bf Top}: results presented in ShapeBM; {\bf Middle}: random results produced by fully-connected DBRN; {\bf Bottom}: random results produced by convolutional DBRN.}
\label{fig:weizmann_compare}
\end{figure}

We also compare our generated images with the architecture introduced by ShapeBM. As shown in Figure~\ref{fig:weizmann_compare}, both models are able to generate reasonable horses. Most of the images generated by ShapeBM share the same posture, though they do have abundant details regarding the legs. Our models, on the other hand, can generate horses with a wider range of variety in general. It is also worth noting that the model of ShapeBM requires layer-wise pretraining, as does deep Boltzmann machine, in order to generate meaningful images, while our DBRN models do not require such procedure and thus can be train directly.

\section{Conclusion}
In this paper, we proposed a new way to compose restricted Boltzmann machine and convolutional restricted Boltzmann machine, forming a deep network to improve its performance in terms of both image generation and feature extraction. We also provided a simple and intuitive training method that jointly optimize all RBMs in the network, which turns out to work well in practice. Our experiments on MNIST and Weizmann Horse datasets show that such composite architectures are good generative models and can extract useful features to facilitate supervised learning task like classification. In the future, it would be interesting to see whether this architecture can be used on real-scaled images and whether it can be generalized to use other Boltzmann machines, \eg deep Boltzmann machine, as its basic unit for each layer.

{\small
\bibliographystyle{ieee}
\bibliography{egbib}
}

\end{document}